\pdfoutput=1
\documentclass[11pt]{article}

\usepackage[]{emnlp2021}

\usepackage{times}
\usepackage{latexsym}

\usepackage[T1]{fontenc}
\usepackage[utf8]{inputenc}

\usepackage{microtype}

\usepackage{multirow}
\usepackage{multicol}
\usepackage{booktabs} % To thicken table lines
\usepackage{graphicx}
\usepackage{amsmath}
\usepackage{amssymb}% http://ctan.org/pkg/amssymb
\usepackage{pifont}% http://ctan.org/pkg/pifont
\DeclareMathOperator*{\argmax}{arg\,max}

\newcommand{\cmark}{\ding{51}}%
\newcommand{\xmark}{\ding{55}}%
\newcommand{\ie}{\textit{i}.\textit{e}.}
\newcommand{\eg}{\textit{e}.\textit{g}.}
\newcommand{\etal}{\textit{et al.}}
\newcommand{\etc}{\textit{etc.}}
\usepackage{bbm}
\usepackage{subfigure}

\title{Relation-aware Video Reading Comprehension for \\Temporal Language Grounding}

\author{Jialin Gao\textsuperscript{\rm 1,2,3$*$} \quad Xin Sun\textsuperscript{\rm 1,2}\thanks{ \quad Jialin Gao and Xin Sun are co-first authors with
equal contributions, supervised by Prof. Xi Zhou in SJTU.} \quad MengMeng Xu\textsuperscript{\rm 3} \quad Xi Zhou\textsuperscript{\rm 1,2} \quad Bernard Ghanem\textsuperscript{\rm 3}\\
\textsuperscript{\rm 1}Cooperative Medianet Innovation Center, Shanghai Jiao Tong University \\
    \textsuperscript{\rm 2} CloudWalk Technology Co., Ltd, China \\
	\textsuperscript{\rm 3} King Abdullah University of Science and Technology \\
	
	\textsuperscript{}\{jialin\_gao, huntersx\}@sjtu.edu.cn,~zhouxi@cloudwalk.cn\\
	\{mengmeng.xu,bernard.ghanem\}@kaust.edu.sa
}

\begin{document}
\maketitle

\begin{abstract}
Temporal language grounding in videos aims to localize the temporal span relevant to the given query sentence. Previous methods treat it either as a boundary regression task or a span extraction task. This paper will formulate temporal language grounding into video reading comprehension and propose a \textit{Relation-aware Network (RaNet)} to address it. This framework aims to select a video moment choice from the predefined answer set with the aid of coarse-and-fine choice-query interaction and choice-choice relation construction. A choice-query interactor is proposed to match the visual and textual information simultaneously in sentence-moment and token-moment levels, leading to a coarse-and-fine cross-modal interaction. Moreover, a novel multi-choice relation constructor is introduced by leveraging graph convolution to capture the dependencies among video moment choices for the best choice selection. Extensive experiments on ActivityNet-Captions, TACoS, and Charades-STA demonstrate the effectiveness of our solution. Codes have been available at \url{https://github.com/Huntersxsx/RaNet}.
\end{abstract}
\section{Introduction}
\begin{figure}
    \centering
    \includegraphics[width=1.0\linewidth]{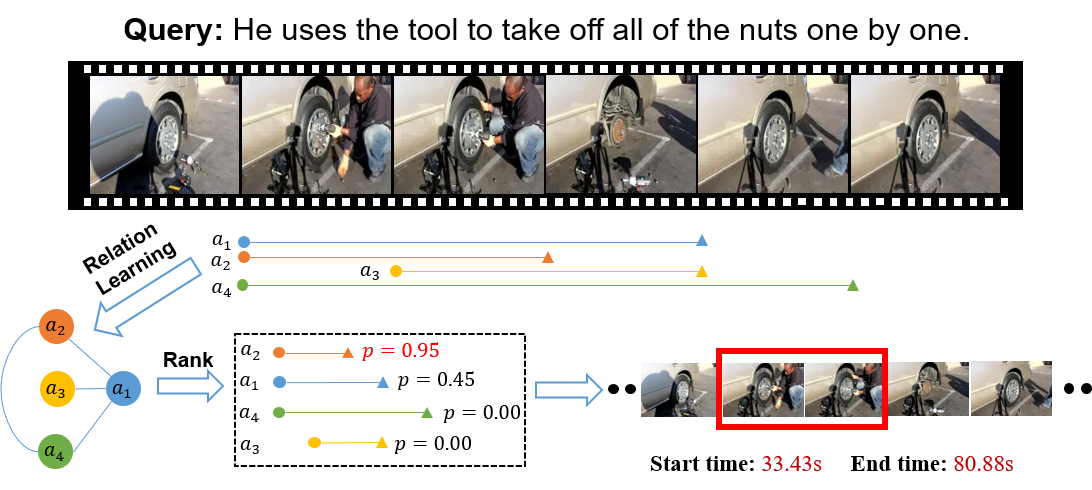}
    \caption{\textbf{An illustration of temporal language grounding in videos based on the relation-aware network.} Given a video and a query sentence, our approach aims to semantically align the query representation with a predefined answer set of video moment candidates ($a_1$,$a_2$,$a_3$ and $a_4$) and then mine the relationships between them  to select the best-matched one.
    }
    \label{fig:motivation}
\end{figure}

Recently, temporal language grounding in videos has become a heated topic in the computer vision, and natural language processing community \cite{gao2017tall,krishna2017dense}. This task requires a machine to localize a temporal moment semantically relevant to a given language query, as shown in Fig.\ref{fig:motivation}. It has also drawn great attention from industry due to its various applications such as video question answering \cite{huang2020location,lei2018tvqa}, video content retrieval \cite{dong2019dual,shao2018find}, and human-computer interaction \cite{zhu2020vision}, \etc

A straightforward paradigm for this task is the proposing-and-ranking pipelines \cite{xu2019multilevel,zhang2020learning,zhang2019man}. They first generate a number of video moment candidates and then rank them according to moment-query similarities. It requires a solution to achieve two key targets simultaneously, which are (1) \textit{semantic visual-language interaction} and (2) \textit{reliable candidate ranking}. The former ensures a satisfying cross-modal matching between video moments and the query, while the latter guarantees the distinction among candidates. For the first target, some previous works \cite{yuan2019find,zhang2020learning,chen2019semantic} resort to the visual clues by modeling moment-sentence or snippet-sentence relations.
% deal with the first target by aligning the linguistic meanings of query sentences with the visual clues from either moment-level or snippet-level relations. 
However, they overlook the linguistic clues from token-level, \ie, token-moment relations, which contain fine-grained linguistic information. 
For the second target, previous solutions \cite{ge2019mac,liu2018attentive,zhang2020span} generate ranking scores by considering different moment candidates separately or constructing moment-level relations in a simple way\cite{zhang2020learning}. Hence, they neglect the temporal and semantic dependencies among candidates. Without this information, it is difficult for previous approaches to distinguish these visually similar moment candidates correctly. 

To this end, we propose a \textit{Relation-aware Network (RaNet)} to address temporal language grounding. In our solution, we formulate this task as Video Reading Comprehension by regarding the video, query, and moment candidates as the text passage, question description, and multi-choice options, respectively. 
% Inspired by the progress in machine reading comprehension \cite{zhang2020dcmn+}, we then propose to explore a finer semantic interaction between visual and textual modalities. 
Unlike previous methods, we exploit a coarse-and-fine interaction, which captures not only sentence-moment relations but also token-moment relations. This interaction can allow our model to construct both sentence-aware and token-aware visual representation for each choice, which is expected to distinguish similar candidates in the visual modality. Moreover, we propose to leverage Graph Convolutional Networks (GCN) \cite{kipf2016semi} for mining the moment-moment relations between candidate choices based on their coarse-and-fine representations. With information exchange in GCNs, our RaNet can learn discriminative features for correctly ranking candidates regardless of their high relevance in visual content.

Similar to the system of multi-choice reading comprehension, our RaNet consists of five components: a modality-wise encoder for visual and textual encoding, a multi-choice generator for answer set generation, a choice-query interactor for cross-modality interaction, a multi-choice relation constructor for relationships mining and an answer ranker for the best-matched choice selection. Our contributions are summarized as three-fold:
(1) We address temporal language grounding by a Relation-aware Network, which formulates this task as a video reading comprehension problem. (2) We exploit the visual and linguistic clues exhaustively, i.e., coarse-and-fine moment-query relations and moment-moment relations, to learn discriminative representations for distinguishing candidates.
% We propose coarse-and-fine moment-query relations and moment-moment to learn discriminative representation for distinguishing candidates, which is enhanced with moment-moment relations.
% design a novel RaNet to select the best-matched video moment candidate corresponding to the given query, where both coarse-and-fine moment-query relations and moment-moment relations are exploited for correct candidate ranking. 
(3) The proposed RaNet outperforms other state-of-the-art methods on three widely-used challenging benchmarks: TACoS, Charades-STA and ActivityNet-Captions, where we improve the grounding performance by a great margin (\ie, 33.54\% \textit{v.s.} 25.32\% of 2D-TAN on TACoS dataset).

% with the query language sentence, the key goal is to find where the semantically related video moment is. That requires the network to align the linguistic meanings of the query sentence with the visual clues from moment candidates. We seek to propose the fusion module that can effectively align visual and linguistic information and aggregate them in a dynamic way.
% multi-hop interaction in single modality
% recasting the temporal language grounding task into multi-choice based span extraction in machine reading comprehension.
% To empower our model with the boundary-aware ability, we combine the boundary information to construct the span-level feature representation with a construction function (addition, concatenation, RoI sampling).
% They cannot model overlapping span 

% \begin{figure}
%     \centering
%     \includegraphics[scale=0.4]{pic/MRCS.png}
%     \caption{The diagram of proposed relation-aware machine reading comprehension system.}
%     \label{fig:ramrc}
% \end{figure}
\section{Related Work}
\textbf{Temporal Language Grounding}. This task was introduced by \cite{anne2017localizing,gao2017tall} to locate relevant moments given a language query. He \etal~\cite{he2019read} and Wang \etal~\cite{wang2019language} used reinforcement learning to solve this problem. Chen \etal~\cite{chen2018temporally} and Ghosh \etal~\cite{ghosh2019excl} proposed to select the boundary frames based on visual-language interaction. Most of recent works \cite{xu2019multilevel,yuan2019find,zhang2020learning,chen2019semantic,zhang2019man} adopted the two-step pipeline to solve this problem.\\
%  Recently,
% (Gao et al. 2017) and (Hendricks et al. 2017) extended the task to more general scenarios. (Gao et al. 2017) proposed to jointly model video clips and text queries using multi-modal operations, then alignment scores and location offsets were predicted based on the multi-model representation.
% (Hendricks et al. 2017) proposed to embed both modalities into a common space and minimize the squared distances. Both (Gao et al. 2017) and (Hendricks et al. 2017) exploited temporal visual contexts for localization. (Wu and Han 2018) integrated multiple interactions between different
% modalities and proposed Multi-modal Circulant Fusion. (Liu et al. 2018b) designed a memory attention network to enhance the visual features.
% The early works treat TLG as a ranking
% task, and rely on multi0modal matching architecture
% to find the best matching video moment for a language query. 
\noindent
\textbf{Visual-Language Interaction.} It is vital for this task to semantically match query sentences and videos. This cross-modality alignment was usually achieved by attention mechanism~\cite{46201} and sequential modeling~\cite{hochreiter1997long,medsker2001recurrent}. Xu \etal~\cite{xu2019multilevel} and Liu \etal~\cite{liu2018attentive} designed soft-attention modules while Hendricks \etal~\cite{anne2017localizing} and Zhang \etal~\cite{zhang2019exploiting} chose the hard counterpart. Some works~\cite{chen2018temporally,ghosh2019excl} attempted to use the property of RNN cells and others went beyond it by dynamic filters~\cite{zhang2019man}, Hadamard product~\cite{zhang2020learning}, QANet~\cite{lu2019debug} and circular matrices~\cite{wu2018multi}. However, these alignments neglect the importance of token-aware visual feature in cross-modal correlating and distinguishing the similar candidates.\\
\noindent
\textbf{Machine Reading Comprehension.} Given the reference document or passage, Machine Reading Comprehension (MRC) requires the machine to answer questions about it \cite{zhang2020machine}. There are two types of the existing MRC variations related to the temporal language grounding in videos, \ie, \textit{span extraction} and \textit{multi-choice}. The former \cite{rajpurkar2016squad} extracts spans from the given passage and has been explored in temporal language grounding task by some previous works \cite{zhang2020span,lu2019debug,ghosh2019excl}. The latter \cite{lai2017race,sun2019dream} aims to find the only correct option in the given candidate choices based on the given passage. We propose to formulate this task from the perspective of multi-choice reading comprehension. Based on this formulation, we focus on the visual-language alignment in a token-moment level. Compared with query-aware context representation in previous solutions, we aim to construct token-aware visual feature for each choice. Inspired by recent advanced attention module \cite{gao2020accurate,9009011}, we mine the relations between multi-choices in an effective and efficient way.

\begin{figure*}[ht]
    \centering
    \includegraphics[width=1.0\linewidth]{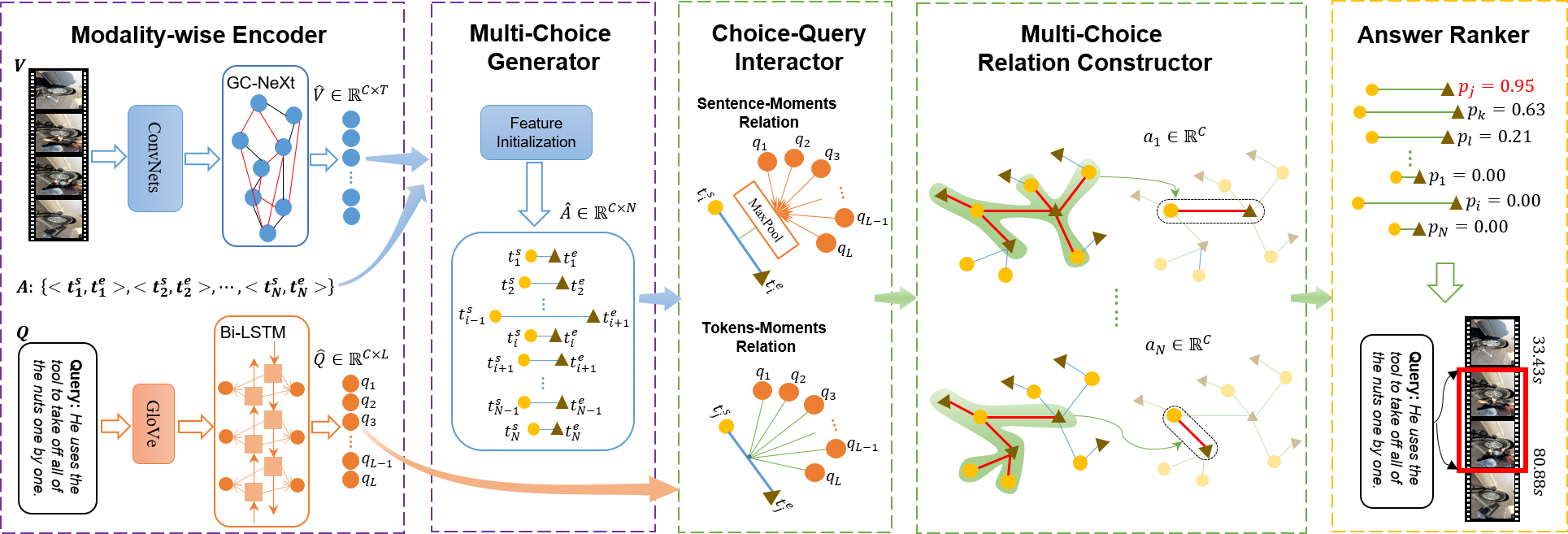}
    \caption{The overview of our proposed RaNet. It consists of modality-wise encoder, multi-choice generator, choice-query interactor, multi-choice relation constructor, and answer ranker. Video and language passages are first embedded in separately branches. Then we initialize the visual representation for each choice $<t_i^s,t_i^e>$ from the video stream. Through choice-query interactor, each choice can capture the sentence-aware and token-aware representation from the query. Afterwards, the relation constructor takes advantage of GCNs to model relationships between choices. Finally, the answer ranker evaluates the probability of being selected for each choice based on the exchanged information from the former module.
    % by the proposed dynamic fusion to learn the cross-modality representations. It is followed by the relation integration module to explore the feature ineraction between moment-query pairs, which aims to predict a better ranking score for video grounding with natural language.
    }
    \label{fig:fw}
\end{figure*}

\section{Methodology}
In this section, we first describe how to cast the temporal language grounding from the perspective of a multi-choice reading comprehension task, which is solved by the proposed Relation-aware Network (RaNet). Then, we introduce the detailed architecture of the RaNet, consisting of five components as shown in Fig.\ref{fig:fw}. Finally, we illustrate the training and inference of our solution.
% Afterwards, these components are deeply integrated to enable our model holistically and exhaustively mining the feature relations  in the training and inference 
% how to construct the relation-aware machine reading comprehension system for better temporal language grounding. As shown in , our framework consists of five main components: modality-wise encoder, multi-choice generator, choice-query interactor, multi-choice relation constructor and answer ranker. These components are deeply integrated and thus enable our model holistically and exhaustively exploring the feature relations both in local (snippets or tokens) and global (choice-query pairs) level.

% \subsection{Notation}
% V: untrimmed video\\
% X: video frames\\
% L: length of video frames\\
% $n_v$: the length of video representations\\
% $n_q$: the length of queries\\
% T: resize the temporal scale to fixed length\\
% Q: a set of language queries\\
% M: the number of the words in a sentence\\
% A: a set of video moment candidates\\
% K: the number of the neighbors\\
% C: the number of channels \\

\subsection{Problem Definition}
\label{pd}
The goal of this task is to answer where is the semantically corresponding video moment given a language query in an untrimmed video. Referring to the forms of MRC, we treat the video $\mathbf{V}$ as a text passage, the query sentence $\mathbf{Q}$ as a question description and provide a set of video moment candidates as a list of answer options $\mathbf{A}$. Based on the given triplet $(\mathbf{V,Q,A})$, temporal language grounding in videos is equivalent to cross-modal MRC, termed as video reading comprehension. 

For each query-video pair, we have one natural language sentence and an associated ground-truth video moment with the start $g^s$ and end $g^e$ boundary. Each language sentence is represented by $\mathbf{Q}=\{q_i\}^{L}_{i=1}$, where $L$ is the number of tokens. The untrimmed video is represented as a sequential snippets $\mathbf{V}=\{v_1, v_2, \cdots, v_{n_v}\} \in \mathcal{R}^{n_v\times C}$ by a pretrained video understanding network, such as C3D~\cite{tran2015learning}, I3D~\cite{carreira2017quo},VGG~\cite{simonyan2014very} \etc. %feature extractor, like a C3D model.
% first denoted as a sequence of frames $\mathbf{X}=\{x_1, x_2, \cdots, x_{n_x}\}$. Through the , it can be , where each $v_i$ contains the $\sigma$ consecutive frames in $\mathbf{X}$ and $n_v = n_x / \sigma$. 

In temporal language grounding, the answer should be a consecutive subsequence (namely time span) of the video passage. For any video moment candidate $(i,j)$, it can be treated as a valid answer if it meets the condition of $0<i<j<n_{v}$. Hence, we follow the fixed-interval sampling strategy in previous work \cite{zhang2020learning} and construct a set of video moment candidates as the answer list $A=\{a_1, \cdots, a_N\}$ with $N$ valid candidates. After these notations, we can recast the temporal language grounding task from the perspective of multi-choice reading comprehension as:
\begin{equation}
    \begin{aligned}
        \argmax_{i}P(a_{i}|(V, Q, A)).\\
        % s.t \quad 0<i<j<v \quad
    \end{aligned}
    \label{eq:se}
\end{equation}

However, different from the traditional multi-choice reading comprehension, previous solutions in temporal language grounding also compare their performance in terms of top-$K$ most matching candidates for each query sentence. For fair comparison, it requires our approach to scores $K$ candidate moments $\{(p_i, t^s_i, t^e_i)\}_{i=1}^K$, where $p_i, t^s_i, t^e_i$ represent the probability of selection, the start, end time of answer $a_i$, respectively. Without additional mention, the video moment and answer/choice are interchangeable in this paper.

\subsection{Architecture}
% First of all, we employ the same feature extractor in the embedding processing as previous methods \cite{zhang2020learning,zhang2020span}. Then, our method starts with a modality-wise encoder to separately encode the content of language query sentence and video. A multi-choice generator is used to initialize the visual feature of each moment choice based on the video representation. Afterwards, a novel choice-query interactor is proposed to calculate the sentence-aware and token-aware visual representation by integrating the language encoding into each initialized moment-level feature. Finally, we explore the relations among the answer candidates by the multi-choice relation constructor, which is followed by an answer ranker.

As shown in Figure~\ref{fig:fw}, we describe the details of each component in our framework as followings:
\noindent
\textbf{Modality-wise Encoder.} This module aims to separately encode the content of language query and video. Each branch aggregates the intra-modality context for each snippet and token.

\textit{$\cdot$ Video Encoding.} We first apply a simple temporal 1D convolution operation to map the input feature sequence to a desired dimension, which is followed by an average pooling layer to reshape the sequence into a desired length $T$. To enrich the multi-hop interaction, we use a graph convolution block called GC-NeXt block \cite{Xu_2020_CVPR}, which aggregates the context from both temporal and semantic neighbors of each snippet $v_i$ and has been proved effective in Temporal Action Localization task. Finally, We get the encoded visual feature as $\hat{\mathbf{V}}\in\mathcal{R}^{C\times T}$

\textit{$\cdot$ Language Encoding.} Each word $q_i$ of $\mathbf{Q}$ is represented with the embedding vector from GloVe 6B 300d \cite{pennington2014glove} to get $\mathbf{Q}\in \mathcal{R}^{300\times L}$. Then we sequentially feed the initialized embeddings into a three-layer Bi-LSTM network to capture semantic information and temporal context. We take the last layer's hidden states as the language representation $\hat{\mathbf{Q}}\in \mathcal{R}^{C\times L}$ for cross-modality fusion with video representation $\hat{\mathbf{V}}$. In addition, the effect of different word embeddings, is also compared in ~\ref{sec:embed}.

The encoded visual and textual features can be formulated as follows:
\begin{equation}
\begin{split}
        \hat{\mathbf{V}} &= VisualEncoder(\mathbf{V})\\
        \hat{\mathbf{Q}} &= LanguageEncoder(\mathbf{Q})\\
\end{split}
\end{equation}

\begin{figure}[t]
    \centering
    \includegraphics[width=1.0\linewidth]{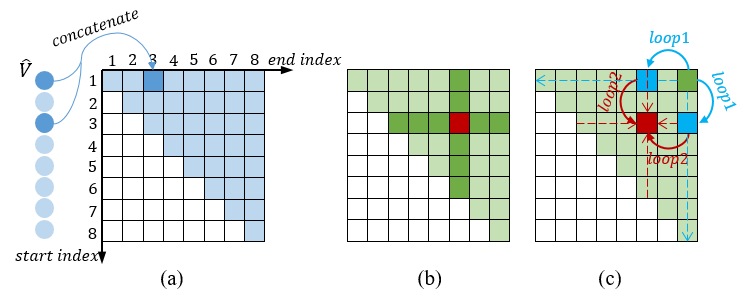}
    \caption{
    (a). Illustration of choice generator and feature initialization. Each block $(i,j)$ is a valid choice when $i<j$, denoted with blue. The $\Psi$ combines the boundary feature for feature initialization of $(1,3)$, the dark blue square.
    (b). An example of all graph edges connected to one choice in our Multi-Choice Relation Constructor. Moments with the same start or end index (dark green) are connected with the illustrative choice (red).
    (c). The information propagation between two unconnected moment choices. For other moments (dark green) that are not connected with target moment (red) but have overlaps, relations can be implicit captured with two loops, namely 2 graph attention layers. 
    }
    \label{fig:gen}
\end{figure}
\noindent
\textbf{Multi-Choice Generator.} As shown in Fig.\ref{fig:gen} (a), the vertical and horizontal axes represent the start and end index of visual sequence. The blocks in the same row have the same start indices, and those in the same column have the same end indices. The white blocks indicate the invalid choices in the left-bottom, where the start boundaries exceed the end boundaries. Therefore, we have the multi-choice options as $A=\{(t^s_i,t^e_i)\}^N_{i=1}$. To capture the visual-language interaction, we should initialize the visual feature for the answer set $A$ so that it can be integrated with the textual features from the language encoder. 
% Based on the visual representation from the video encoder, we adopt a construction function $\Phi$ for feature initialization.
% Inspired by previous work , we combine the boundary information to 
% For the output of video stream, we perform a fixed-interval ($\tau$) sampling from each temporal position $i$ to the end of the video sequence $F_v$ so that cut the whole video into $n$ candidate moments. We reshape it into a 2D map, illustrated in Fig.\ref{fig:cc}, where one dimension indicates the starting time of a moment and the other indicates the end time. Due to the time order constraint, it is a valid moment candidate $(i,j)$ if and only if $0<\leg i<j\leq T$, shown as the green blocks. 
To ensure the boundary-aware ability inspired by \cite{wang2020temporally}, the initialization method $\Psi$ combines boundary information, \ie, $\hat{v}_{t^s_i}$ and $\hat{v}_{t^e_i}$ in $\hat{\mathbf{V}}$, to construct the moment-level feature representation for each choice $a_i$. The initialized feature representation can be written as:
\begin{equation}
% \begin{split}
        \mathbf{F}_A = \Psi(\hat{\mathbf{V}}, A),
% \end{split}
\end{equation}
where $\Psi$ is the concatenation of $\hat{v}_{t^s_i}$ and $\hat{v}_{t^e_i}$, $A$ is the answer set and $\mathbf{F}_A\in\mathcal{R}^{2C\times N}$. We also explore the effect of different $\Psi$ on grounding performance in ~\ref{sec:initial}.\\
% as addition, pooling, concatenation operation in the section of ablation study for comparison with the RoI sampling counterpart.
\noindent
\textbf{Choice-Query Interactor.} 
% The modality-wise encoder aims to capture the rich intra-modality context between each snippet or token.
As shown in Figure~\ref{fig:fw}, this module explores the inter-modality context for visual-language interaction. Unlike previous methods \cite{zhang2020learning,zhang2020span,Zeng_2020_CVPR}, we propose a \textit{coarse-and-fine} cross-modal interaction. We integrate the initialized features $\mathbf{F}_A$ with the query both in the sentence-level and token-level. The former can be obtained by a simple Hadamard product and an normalization as:
\begin{equation}
    \mathbf{F}_1 = \Vert \varphi(\hat{\mathbf{Q}}) \odot Conv(\mathbf{F}_A) \Vert_F,
\end{equation}
where $\varphi$ is the aggregation function for a global representation of $\hat{\mathbf{Q}}$ and we set it as max-pooling, $\odot$ is element-wise multiplication, and $\Vert \cdot \Vert_F$ denotes the Frobenius normalization.

To ensure the token-aware visual feature for each choice $a_i$, we adopt attention mechanism to learn the token-moment relations between each choice and the query. Firstly, we adopt a 1D convolution layer to project the visual and textual features to a common space and then calculate their semantic similarities, which depict the relationships $\mathbf{R}\in \mathcal{R}^{N\times L}$ between $N$ candidates and $L$ tokens. Secondly, we generate query-related feature for each candidate based on the relationships $\mathbf{R}$. Finally, we integrate these two features of candidates for token-aware visual representation. 
\begin{equation}
\begin{split}
    \mathbf{R} &= Conv(\mathbf{F}_A)^T \otimes Conv(\hat{\mathbf{Q}}) \\
    \mathbf{F}_2 &= Conv(\mathbf{F}_A) \odot ( Conv(\hat{\mathbf{Q}}) \otimes \mathbf{R}^T),
\end{split} 
\end{equation}
where $T$ denotes the matrix transpose, $\odot$ and $\otimes$ are element-wise and matrix multiplications, respectively. We add the sentence-aware $\mathbf{F}_1$ and token-aware features $\mathbf{F}_2$ as the output of this module $\hat{\mathbf{F}}_A$.

\noindent
\textbf{Multi-choice Relation Constructor.} 
% Previous methods MAN \cite{zhang2019man}, 2D-TAN \cite{zhang2020learning} also considered moment-wise temporal relations. MAN devised an iterative graph adjustment network to learn dense adjacency matrice, while our implementation is similar to 2D-TAN, which utilized a sparse sampling strategy. However, 2D-TAN simply stacked several convolution layers, which is less effective and computationally expensive, our relation constructor considers more boundary information and is implemented in a faster way.
In order to explore the relation between multi-choices, we propose this module to aggregate the information from the overlapping moment candidates by GCNs. 
% They only use some convolutional layer to predict the ranking score of each moment separately. However, we model the relation between them and capture more moment-level contextual information, which is expected for a better ranking score. 
% A direct way of modeling the relation is employing the fully-connected graph network. 
Previous methods MAN \cite{zhang2019man}, 2D-TAN \cite{zhang2020learning} also considered moment-wise temporal relations, while both of them have two drawbacks: expensive computations and the noise from unnecessary relations. Inspired by CCNet \cite{9009011}, which proposed a sparsely-connected graph attention module to collect contextual information in horizontal and vertical directions, we propose a Graph ATtention layer (GAT) to construct the relation between moment candidates that have high temporal overlaps with each other. 
% For each moment $(i, j)$, we only consider the relations from the moment candidate, which has the same start time $i$ or end time $j$. Therefore, the relations can be learned in a criss-cross way, shown in Fig.\ref{fig:cc}.

% We construct a graph to model the relation between moment candidates, and apply Graph Attention (GAT) layer to aggregate the information from related moment candidates.
Concretely, we take each answer candidate $a_i= (t^s_i,t^e_i)$ as a graph node, and there is a graph edge connecting two candidate choices $a_i$, $a_j$ if they share the same start/end time spot, \eg, $t^s_i=t^s_j$ or $t^e_i=t^e_j$. An example is shown in Figure~\ref{fig:gen} (b), where neighbors of the target moment choice (the red one) is denoted as dark green in a criss-cross shape. As shown in Figure~\ref{fig:gen} (c), our model is also able to achieve the information propagation between two unconnected moment choices. For other moments (dark green) that are not connected with the target moment (red) but have overlapped, their relations can be implicitly captured with two loops, namely two graph attention layers. We can guarantee the message passing between the dark green moment and cyan moments in the first loop. And then, in the second loop, we can construct relations between cyan moments and target moment, where the information from the dark green moment is finally propagated to the red moment.

Given the choice-query $\hat{\mathbf{F}}_A\in \mathbb{R}^{C\times N}$, there are $N$ nodes and approximately $2TN$ edges in the graph. 
A GAT layer inpsired by
%\cite{velivckovic2017graph} is applied on the graph: for each node, we compute attention weights of its neighbours, and average the neighbours feature with the weights.
\cite{9009011} is applied on the graph: for each moment, we compute attention weights of its neighbours in a criss-cross path, and average the features with the weights. The output of the GAT layer can be formulated as:
\begin{equation}
    \hat{\mathbf{F}}_A^* = Conv(GAT(Conv(\hat{\mathbf{F}}_A), \hat{A}))\\
\end{equation}
where $\hat{A}$ is the adjacency matrix of the graph to determine the connections between two moment choices, defined by the predefined answer set $A$.
% where $\hat{\mathbf{F}}_A^* \in \mathcal{R}^{C\times N}$ and $\hat{A}$ is the adjacency matrix used to determine the connections between two moment choice. $\hat{A}$ is defined by the predefined answer set $A$.
% Then we concatenate the averaged feature with the node feature to aggregate more contextual information in the neighbourhood. 
\noindent
\textbf{Answer Ranker.} Since we have captured the relationship between multi-choices by GCNs, we adopt this answer ranker to predict the ranking score of each answer candidate $a_i$ for selecting the best-matched one. 
This ranker takes the query-aware feature $\hat{\mathbf{F}}_A$ and
relation-aware feature $\hat{\mathbf{F}}_A^*$ as input 
and concatenate them (denoted as $\Vert$) to aggregate more contextual information. 
After that, we employ a convolution layer to generate the probability $P_A$ of being selected for $a_i$ in the predefined answer set $A$. The output can be computed as:
\begin{equation}
    P_A = \sigma(Conv( \hat{\mathbf{F}}_A^* \Vert \hat{\mathbf{F}}_A )),
\end{equation}
where $\sigma$ represents the \textit{sigmoid} activation function.

\subsection{Training and Inference}
Following \cite{zhang2020learning}, we first calculate the Intersection-over-Union (IoU) between the answer set $A$ and ground-truth annotation $(g^s,g^e)$ and then rescale them by two thresholds $\theta_{min}$ and $\theta_{max}$, which can be written as:
\begin{equation}
   g_i = 
    \left\{
    \begin{array}{cc}
         0 & \theta_i \leq \theta_{min} \\
         \frac{\theta_i - \theta_{min}}{\theta_{max} - \theta_{min}} &  \theta_{min} < \theta_i < \theta_{max}  \\
         1 & \theta_i \geq \theta_{max}
    \end{array}
    \right.
\end{equation}
where $g_i$ and $\theta_i$ are the supervision label and corresponding IoU between $a_i$ and ground-truth respectively. Hence, the total training loss function of our RaNet is:
\begin{equation}
    \mathcal{L} = -\frac{1}{N} \Sigma_{i=1}^{N}(g_i\log p_i + (1 - g_i)\log(1 - p_i)),
\end{equation}
where $p_i$ is the output score in $P_A$ for the answer choice $a_i$ and $N$ is the number of multi-choices. In the inference stage, we rank all the answer options in $A$ according to the probability in $P_A$.

\section{Experiments}

\begin{table}[]
    \centering
    % \footnotesize
    \begin{tabular}{ccccc}
    \toprule
    \multirow{2}{*}{\textbf{Methods}} & \multicolumn{2}{c}{\textbf{Rank1@}}  & \multicolumn{2}{c}{\textbf{Rank5@}} \\
    & \textbf{0.3} & \textbf{0.5} & \textbf{0.3} & \textbf{0.5} \\
    \midrule
    MCN  &  - & 5.58 & - & 10.33\\
    CTRL & 18.32 & 13.30 & 36.69 & 25.42 \\
     % MCF & 18.64 & 12.53  & 37.13 &24.73\\
    ACRN &19.52 &14.62 &34.97 &24.88\\
    ROLE &15.38& 9.94 &31.17 &20.13\\
    TGN & 21.77 & 18.9 & 39.06 & 31.02 \\
    % VAL &19.76 &14.74  &38.55 &26.52\\
    % SAP & - &18.24 & - &28.11\\
    ABLR  &19.50 &9.40  & -&  -\\
    SM-Rl &20.25 &15.95 &38.47& 27.84\\
    CMIN &24.64 &18.05  &38.46 &27.02\\
    QSPN  &20.15 &15.23 &36.72 &25.30\\
    ACL-K &24.17 &20.01 &42.15 &30.66\\
    % SLTA &17.07 &11.92  &32.90& 20.86\\
    % TripNet &23.95 &19.17 & - &-\\
    2D-TAN  &\underline{37.29} &\underline{25.32}  &\underline{57.81} &\underline{45.04}\\
    DRN  & - & 23.17 & - & 33.36 \\
    DEBUG & 23.45 & 11.72 & - & -\\
    VSLNet & 29.61 & 24.27  & - & -\\
    \midrule
    Ours  & \textbf{43.34} & \textbf{33.54}  & \textbf{67.33} & \textbf{55.09} \\ 
    \bottomrule
    \end{tabular}
    \caption{Performance comparison on TACoS. All results are reported in percentage (\%).} % To save space, we put the whole table in the Appendix. }
    \label{tab:tacos}
\end{table}

To evaluate the effectiveness of the proposed approach, we conduct extensive experiments on three public challenging datasets: TACoS \cite{regneri2013grounding}, ActivityNet Captions \cite{krishna2017dense} and Charades-STA \cite{sigurdsson2016hollywood}. 
% In this section, we begin with detailed introduction on these datasets and then compares the performance of our method with other state-of-the-art approaches. Finally, we perform complete ablation studies to evaluate the effect of the components in our model.

\subsection{Dataset}
\textbf{TACoS.}
 It consists of 127 videos, which contain different activities that happened in the kitchen. We follow the convention in \cite{gao2017tall}, where the training, validation, and testing contain $10,146$, $4,589$, and $4,083$ query-video pairs.\\
\textbf{Charade-STA.} It is extended by \cite{gao2017tall} with language descriptions leading to $12,408$ and $3,720$ query-video pairs for training and testing.\\
% It contains 9, 848 videos of daily indoors
% activities. It is originally designed for action recognition and localization. Gao et al. extend the temporal annotation (i.e. labeling the start and end time of moments) of this dataset  and name it as Charades-STA. Charades-STA contains  moment sentence pairs in training set and 3, 720 pairs in testing set.\\
\textbf{ActivityNet Captions.} It is introduced into the temporal language grounding task recently.
% It consists of 19, 209 videos, whose content are diverse and open. It is originally designed for video captioning task, and recently introduced into the task
% of moment localization with natural language, since these two tasks are reversible. 
Following the setting in CMIN \cite{lin2020moment}, we use val\_1 as validation set and val\_2 as testing set, which have $37,417$, $17,505$, and $17,031$ query-video pairs for training, validation, and testing, respectively.

% \subsection{Evaluation Metric}
% Following the setting as previous work (TALL), we evaluate our model by computing \textit{Rank n@$\mu$}. It is defined as the percentage of language queries having at least one correct moment retrieval in the top-$n$ retrieved moments. A retrieved moment is correct when its Intersection over Union (IoU) with the ground truth is larger than $\mu$. There are specific settings of n and m for different datasets. Specifically, we report the results as $n\in \{1,5\}$ with $\mu\in\{0.5, 0.7\}$ for Charades-STA dataset, $n\in \{1,5\}$ with  $\mu\in\{0.5, 0.7\}$for ActivityNet Captions dataset, and $n\in \{1,5\}$ with m $\mu\in\{0.1, 0.3, 0.5\}$ for TACoS dataset.

\subsection{Implementation Details} 

\noindent
\textbf{Evaluation metric.} Following Gao \etal~\cite{gao2017tall}, we compute the \textit{Rank k@$\mu$} for a fair comparison. It denotes the percentage of testing samples that have at least one correct answer in the top-K choices. A selected choice $a_i$ is correct when its IoU $\theta_i$ with the ground-truth is larger than the threshold $\mu$; otherwise, the choice is wrong. Specifically, we set $k\in\{1,5\}$ and $\mu\in\{0.3, 0.5\}$ for TACoS and $\mu\in\{0.5, 0.7\}$ for the other two.

\noindent
\textbf{Feature Extractor.} We follow the \cite{zhang2019man,lin2020moment} and adopt the same extractor, \eg, VGG \cite{simonyan2014very} feature for Charades and C3D \cite{tran2015learning} for other two. We also use the I3D \cite{carreira2017quo} feature to make comparison with \cite{ghosh2019excl,zhang2020span,Zeng_2020_CVPR} on Charades. For word embedding, we use the pre-trained GloVe 6B 300d \cite{pennington2014glove} as previous solutions \cite{ge2019mac,chen2018temporally}.

\noindent
\textbf{Architecture settings.} In all experiments, we set the hidden units of Bi-LSTM as 256 and the number of reshaped snippet $T$ is defined as 128 for TACoS, 64 for ActivityNet Captions and 16 for Charades-STA. The dimension $C$ of channels is 512. We adopt 2 GAT layers for all benchmarks and position embedding is used in ActivityNet Caption as \cite{Zeng_2020_CVPR}.

\noindent
\textbf{Training settings.} We adopt the Adam optimizer with learning rate of $1\times 10^{-3}$, batch size of 32, and training epoch of 15. Following \cite{zhang2020learning}, thresholds $\theta_{min}$ and $\theta_{max}$ are set to 0.5 and 1.0 for Charades-STA and ActivityNet Captions, while 0.3 and 0.7 for TACoS.

\subsection{Comparison with State-of-the-arts}
Our RaNet is compared with recent published state-of-the-art methods: VLSNet \cite{zhang2020span}, 2D-TAN \cite{zhang2020learning}, DRN \cite{Zeng_2020_CVPR}, CMIN \cite{lin2020moment}, DEBUG \cite{lu2019debug}, QSPN \cite{xu2019multilevel}, MAN \cite{zhang2019man}, ExCL \cite{ghosh2019excl}, CTRL \cite{gao2017tall}, \etc. The top-2 performance values are highlighted by bold and underline, respectively.

\noindent
\textbf{TACoS.} Table \ref{tab:tacos} summarizes performance comparison of different methods on the test split. Our RaNet outperforms all the competitive methods with clear margins and reports the highest scores for all IoU thresholds. Compared with the previous best method 2D-TAN, our model achieves 6\%  absolute improvement at least across all evaluation settings in terms of \textit{Rank 1@$\mu$}, \ie, 8.22\% for $\mu=0.5$. For evaluation metric of \textit{Rank 5@$\mu$}, we even reach around 10\% absolute improvement. It is worth noting that we exceed VSLNet by 9.27\% and 13.73\% in terms of \textit{Rank 1@$\mu=0.5$, $\mu=0.3$} respectively, which also formulates this task from the perspective of MRC. 

\noindent
\textbf{Charades-STA.} 
We evaluate our method both on VGG and I3D features used in previous works for fair comparison. Our approach reaches the highest score in terms of \textit{Rank 1} no matter which kind of feature is adopted as illustrated in Table \ref{tab:sta}. For VGG feature, we improve the performance from 23.68\% in DRN to 26.83\% in terms of \textit{Rank 1@$\mu=0.7$}. By adopting the stronger I3D feature, our method also exceeds VSLNet in terms of \textit{Rank 1@$\mu=\{0.5, 0.7\}$} (\textit{i.e.}, 60.40\% vs. 54.19\% and 39.65\% vs. 35.22\%).

\begin{table}[]
    \centering
    \begin{tabular}{ccccc}
    \toprule
    \multirow{2}{*}{\textbf{Methods}} &\multicolumn{2}{c}{\textbf{Rank1@}}& \multicolumn{2}{c}{\textbf{Rank5@}}\\
    & \textbf{0.5} & \textbf{0.7} & \textbf{0.5} & \textbf{0.7} \\
    % \midrule
    % \multicolumn{5}{|c|}{C3D}\\
    % \midrule
    % VAL& 23.12 &9.16& 61.26 & 27.98  \\
    % TripNet& \underline{38.29} &16.07&-&-\\
    % ROLE& 21.74 &7.82 & 70.37 & 30.06\\
    % ACL-K& 30.48 &12.20 & 64.84& 35.13\\
    % QSPN& 35.60 &15.80 & 79.40& 45.40 \\
    % DEBUG & 37.39 & \underline{17.69} & - &- \\
    % CTRL&23.63 &8.89&58.92 & 29.52 \\
    % DRN &45.40 & 26.40 & 88.01 & 55.38 \\
    % Ours&\textbf{xxx} & \textbf{xxx}& \textbf{xxx} & \textbf{xxx}\\ 
    \midrule
    \multicolumn{5}{c}{\textbf{VGG}}\\
    \midrule
    MCN & 17.46 & 8.01 &48.22  & 26.73 \\
    CTRL&23.63 &8.89&58.92 & 29.52 \\
    ABLR& 24.36 &9.01 & -& -  \\
    % SAP& 27.42 &13.36 & 66.37 & 38.15 \\
    QSPN& 35.60 &15.80 & 79.40& 45.40 \\
    ACL-K& 30.48 &12.20 & 64.84& 35.13\\
    DEBUG & 37.39 & 17.69 & - &- \\
    MAN& 41.24 &20.54  & 83.21& 51.85 \\
    2D-TAN& 39.70 &23.31 & 80.32 & 51.26\\
    DRN&\underline{42.90} &\underline{23.68}  & \textbf{87.80} & \textbf{54.87} \\
    % Ours&VGG & \textbf{} & \textbf{44.54} & \textbf{27.53}\\
    Ours& \textbf{43.87} & \textbf{26.83} & \underline{86.67}  & \underline{54.22} \\
    \midrule
    \multicolumn{5}{c}{\textbf{I3D}}\\
    \midrule
    ExCL& 44.10 & 22.40 & - &-\\
    % VSLNet&I3D & 64.30 & 47.31 & 30.19 \\
    VSLNet& \underline{54.19} & \underline{35.22} & -& -  \\
    DRN  & 53.09 & 31.50 & 89.06 & 60.05 \\
    Ours& \textbf{60.40} & \textbf{39.65} & \textbf{89.57} & \textbf{64.54}\\
    \bottomrule
    \end{tabular}
    \caption{Performance comparison on Charades-STA. All results are reported in percentage (\%).} % To save space, we put the whole table in the Appendix.}
    \label{tab:sta}
\end{table}

\noindent
\textbf{ActivityNet-Captions.} In Table \ref{tab:anet}, we compare our model with other competitive methods. Our model achieves the highest scores over all IoU thresholds in the evaluation except the result of \textit{Rank 5@$\mu=0.5$}. Particularly, our model outperforms the previous best method (\textit{i.e.}, 2D-TAN) by around 1.29\% absolute improvement, in terms of \textit{Rank 1@$\mu=0.7$}. Due to the same sampling strategy for moment candidates, this improvement is mostly attributed to the token-aware visual representation and the relationships mining between multi-choices. 
%It demonstrates that our RaNet can distinguish the visually related moment candidates better than 2D-TAN.

% \begin{table}[]
%     \centering
%     \begin{tabular}{|c|cc|cc|}
%     \toprule
%     \multirow{2}{*}{} & \multicolumn{2}{c|}{Rank1@}  & \multicolumn{2}{c|}{Rank5@} \\
%     & 0.5 & 0.7 & 0.5 & 0.7 \\
%     \midrule
%     MCN & 21.36 & 6.43 & 53.23 & 29.70\\
%     CTRL & 29.01 & 10.34 & 59.17 & 37.54 \\
%     TGN & 27.93 & - & 44.20 & - \\
%     ACRN & 31.67 & 11.25 & 60.34 & 38.57 \\
%     CMIN & 44.62 & 24.48 & 69.66 & 52.96 \\
%     QSPN & 33.26 & 13.43 & 62.39 & 40.78 \\
%     ABLR & 36.79 & - & - & - \\
%     TripNet & 32.19 & 13.93 & - & - \\
%     PMI & 38.28 & 17.83 & - & - \\
%     2D-TAN & 44.05 & \underline{27.38} & 76.65 & \underline{62.26} \\
%     VSLNet & 43.22 & 26.16 & - & - \\
%     DRN & \underline{45.45} & 24.39 & \underline{77.97} & 50.30 \\
%     ExCL & 42.7 & 24.1 & - & - \\
%     DEBUG & 39.72 & - & - & - \\
%     \midrule
%     Ours & \textbf{46.23} & \textbf{29.37} & 76.09 & \textbf{62.33} \\ 
%     \bottomrule
%     \end{tabular}
%     \caption{Performance comparison on ActivityNet Captions. All results are reported in percentage (\%)}
%     \label{tab:anet}
% \end{table}
\begin{table}[]
    \centering
    \begin{tabular}{ccccc}
    \toprule
    \multirow{2}{*}{\textbf{Methods}} & \multicolumn{2}{c}{\textbf{Rank1@}}  & \multicolumn{2}{c}{\textbf{Rank5@}} \\
    & \textbf{0.5} & \textbf{0.7} & \textbf{0.5} & \textbf{0.7} \\
    \midrule
    MCN & 21.36 & 6.43 & 53.23 & 29.70\\
    CTRL & 29.01 & 10.34 & 59.17 & 37.54 \\
    ACRN & 31.67 & 11.25 & 60.34 & 38.57 \\
    TGN & 27.93 & - & 44.20 & - \\
    QSPN & 33.26 & 13.43 & 62.39 & 40.78 \\
    ExCL & 42.7 & 24.1 & - & - \\
    CMIN & 44.62 & 24.48 & 69.66 & 52.96 \\
    
    ABLR & 36.79 & - & - & - \\
    % TripNet & 32.19 & 13.93 & - & - \\
    % PMI & 38.28 & 17.83 & - & - \\
    DEBUG & 39.72 & - & - & - \\
    2D-TAN & 44.05 & \underline{27.38} & 76.65 & \underline{62.26} \\
    DRN & \underline{45.45} & 24.39 & \textbf{77.97} & 50.30 \\
    VSLNet & 43.22 & 26.16 & - & - \\

    \midrule
    Ours & \textbf{45.59} & \textbf{28.67} & \underline{75.93} & \textbf{62.97} \\ 
    \bottomrule
    \end{tabular}
    \caption{Performance comparison on ActivityNet Captions. All results are reported in percentage (\%).} % To save space, we put the whole table in the Appendix.}
    \label{tab:anet}
\end{table}

\begin{table}[t]
	\centering
	\scalebox{0.95}{
	\setlength{\tabcolsep}{2pt}
	\begin{tabular}{cccccc}
    \toprule
     \multirow{2}{*}{\textbf{Datasets}}& \multicolumn{3}{c}{\textbf{Components}} & \multicolumn{2}{c}{\textbf{Rank1@}}\\
      & $\mathbf{F}_1$ & $\mathbf{F}_2$ & $\mathbf{R}$ & \textbf{0.3} & \textbf{0.5}\\
    \midrule
    %  &\checkmark &\checkmark &\checkmark &\checkmark &\checkmark &\checkmark \\
    \multirow{6}{*}{\textbf{TACoS}}  & \cmark&\xmark&\xmark& 40.99 & 28.54\\
                            & \xmark&\cmark&\xmark&  41.26 & 29.22 \\
                            & \cmark&\cmark&\xmark& 41.51 & 29.64\\
                            & \cmark&\xmark&\cmark& 42.26 & 32.04\\
                            & \cmark&\cmark&\cmark&\textbf{43.34} & \textbf{33.54}\\
    \midrule
    { } & $\mathbf{F}_1$ & $\mathbf{F}_2$ & $\mathbf{R}$ & \textbf{0.5} & \textbf{0.7} \\
    \midrule
    \multirow{4}{*}{\textbf{Charades-STA}}&\cmark&\xmark&\xmark&43.06&24.70\\
                            &\xmark&\cmark&\xmark& 42.72&24.33\\
                            & \cmark&\cmark&\xmark&42.10&24.78\\
                            &\cmark& \xmark&\cmark&43.60&25.30\\
                            & \cmark&\cmark&\cmark&\textbf{43.87}&\textbf{26.83}\\
    % \midrule
    % \multirow{4}{*}{ANet-Captions}&\xmark&\xmark&\\
    %                         & \cmark&\xmark&\\
    %                         & \xmark&\cmark&\\
    %                         & \cmark&\cmark&\\
    \bottomrule
	\end{tabular}%
	}
	\vspace{1 ex}
	\caption{Effectiveness of each component in our proposed approach on TACoS and Charades-STA, measured by \textit{Rank 1@$\mu \in\{0.3, 0.5, 0.7\}$}. VGG features are used in Charades-STA. \cmark \ and \xmark \ denote the net with and without that component, respectively.}
	\label{tab:ablation-components}%
	
\end{table}%

% \begin{table}[t]
%     \centering
%     \begin{tabular}{ccccc}
%     \toprule
%     \multirow{2}{*}{\textbf{Operators}} & \multicolumn{2}{c}{\textbf{Rank1@}}  & \multicolumn{2}{c}{\textbf{Rank5@}} \\
%          & \textbf{0.3} & \textbf{0.5} &\textbf{0.3} & \textbf{0.5} \\
%     \midrule
%     % \textbf{Concatenation} & \textbf{43.09} & \textbf{32.24} & \textbf{68.71} & \textbf{55.51} \\
%     % \textbf{Addition} &  42.04& 31.22 & 63.86 & 51.24 \\
%     % \textbf{Pooling} &  40.61& 29.62 & 64.58 & 52.79 \\
%     % \textbf{Sampling}&  41.33& 30.82 & 65.53 & 53.54 \\
%     Pooling &  40.61& 29.62 & 64.58 & 52.79 \\
%     Sampling&  41.33& 30.82 & 65.53 & 53.54 \\
%     Addition &  42.04& 31.22 & 63.86 & 51.24 \\
%     Concatenation & \textbf{43.09} & \textbf{32.24} & \textbf{68.71} & \textbf{55.51} \\
%     \bottomrule
%     \end{tabular}
%     \caption{Effectiveness of different operators used in Multi-Choice Generator on TACoS, measured by \textit{Rank 1@$\mu \in\{0.3, 0.5\}$} and \textit{Rank 5@$\mu \in\{0.3, 0.5\}$}.}
%     \label{tab:abs_tacos}
% \end{table}

\subsection{Ablation Study}
% In this section, we conduct extensive experiments to investigate the specific effect of each component of the proposed RaNet.

% \subsubsection{Semantic Interaction and Relations Construction}
\subsubsection{Effectiveness of Network Components}
We perform complete and in-depth studies on the effectiveness of our choice-query interactor and multi-choice relation constructor based on the TACoS and Charades-STA datasets. On each dataset, we conduct five comparison experiments for evaluation. 
% First, we replace both our alignment and relation module with simple convolution layer, denoted as removing the $\mathbf{F}_2$ and $\mathbf{R}$ but remaining the $\mathbf{F}_1$. 
First, we remove $\mathbf{F}_2$ and $\mathbf{R}$ to explore the \textit{RaNet-base} model, compared with only using $\mathbf{F}_2$.
Then, we integrate the interaction and relation modules into our third and forth experiments respectively. Finally, we show the best performance achieved by our proposed approach. Table \ref{tab:ablation-components} summarizes the grounding results in terms of \textit{Rank 1@$\mu \in\{0.3, 0.5, 0.7\}$}. Without the interaction and relation modules, our framework can achieve 40.99\% and 28.54\% for $\mu=0.3$ and $0.5$ respectively. It outperforms the previous best method 2D-TAN, indicating the power of our modality-wise encoder. When we consider the token-aware visual representation, our framework can bring significant improvements on both datasets. Improvements can also be observed when adding relation module. These results demonstrate the effectiveness of our RaNet on temporal language grounding.

\begin{figure*}[t]
    \centering
    \includegraphics[width=1.0\linewidth]{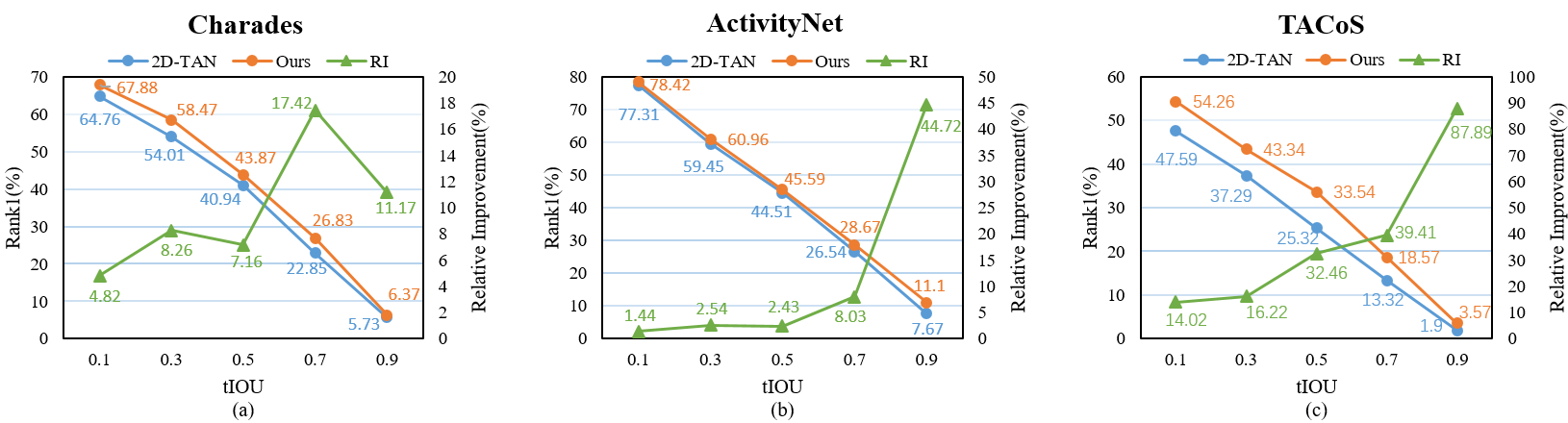}
    \caption{Detailed comparison across different IoUs on three benchmarks in terms of \textit{Rank 1}}
    \label{fig:iou}
\end{figure*}

\subsubsection{Improvement on different IoUs}
To have a better understanding of our approach, we illustrate the performance gain achieved on three datasets in terms of different $\mu\in(0,1)$ with previous best method, 2D-TAN, as shown in Figure~\ref{fig:iou}. This figure visualizes the detailed comparison between our model and 2D-TAN, which shows that our approach can continuously improve the performance, especially for higher IoUs (\ie, $\mu>0.7$ ). It is observed that the value of relative improvement increases along with the increasing IoU on TACoS and ActivityNet Captions datasets.

\subsubsection{Feature Initialization Functions}
\label{sec:initial}
 We conduct experiments to reveal the effect of different feature initialization functions. For a moment candidate $(t^s_i, t^e_i)$, it has the corresponding feature sequence $Y$ from $\hat{\mathbf{V}}$ denoted as $Y=\{\hat{v}_k\}_{k={t^s_i}}^{t^e_i}$. We explore four types of operators (\ie, pooling, sampling, concatenation and addition) in the multi-choice generator. The first two consider all the information of the region within the temporal span of the candidate. Pooling operator focuses on the statistic characteristic and the sampling serves as weight average operator. On the contrary, the last two only consider the boundary information ($v_{t^s_i}$ and $v_{t^e_i}$) of a moment candidate, which expect the cross-modal interaction to be boundary sensitive. Table \ref{tab:init_tacos} reports the performance of different operators on TACoS dataset. It is observed that concatenation operator achieves the highest score across all the evaluation criterion, which indicates boundary sensitive operators have better performance than the statistical operators.
\begin{table}[t]
    \centering
    \begin{tabular}{ccccc}
    \toprule
    \multirow{2}{*}{\textbf{$\Psi$}} & \multicolumn{2}{c}{\textbf{Rank1@}}  & \multicolumn{2}{c}{\textbf{Rank5@}} \\
         & \textbf{0.3} & \textbf{0.5} &\textbf{0.3} & \textbf{0.5} \\
    \midrule
    Pooling &  38.84& 29.29 & 63.31 & 49.86 \\
    Sampling&  41.33& 30.82 & 65.58 & 54.69 \\
    Addition &  42.69& 31.59 & 64.98 & 54.36 \\
    Concatenation & \textbf{43.34} & \textbf{33.54} & \textbf{67.33} & \textbf{55.09} \\
    \bottomrule
    \end{tabular}
    \caption{Effectiveness of different operators used in Multi-Choice Generator on TACoS, measured by \textit{Rank 1@$\mu \in\{0.3, 0.5\}$} and \textit{Rank 5@$\mu \in\{0.3, 0.5\}$}.}
    \label{tab:init_tacos}
\end{table}

\subsubsection{Word Embeddings Comparison}
\label{sec:embed}
To further explore the effect of different textual features, we also conduct experiments on four pre-trained word embeddings (\ie, GloVe 6B, GloVe 42B, GloVe 840B and BERT$_{Base}$). GloVe \cite{pennington2014glove} is an unsupervised learning algorithm for obtaining vector representations for words, which has some common word vectors trained on different corpora of varying sizes. % GloVe 6B is trained on Wikipedia 2014 and Gigaword 5 which totally have 6 billion tokens; GloVe 42B is trained on 42 billion tokens of web data from Common Crawl, and GloVe 840B is a much larger vector trained on a corpus containing 840 billion tokens.
BERT \cite{devlin-etal-2019-bert} is a language representation model considering bidirectional context, which achieved SOTA performance on many NLP tasks.
All the GloVe vectors have 300 dimensions whereas BERT$_{Base}$ is a 768-dimensional vector. Table \ref{tab:embedding} compares the performance of these four pre-trained word embeddings on TACoS dataset. From the results we can see that better word embeddings (\ie BERT) tend to have better performance, indicating us pay more attention to textual features encoding. All the models in our paper use  concatenation as multi-choice feature  initialization  function and GloVe 6B word vectors for word embeddings initialization  if not specified.
% \begin{table}[t]
%     \centering
%     \begin{tabular}{ccccc}
%     \toprule
%     \multirow{2}{*}{\textbf{Methods}} & \multicolumn{2}{c}{\textbf{Rank1@}}  & \multicolumn{2}{c}{\textbf{Rank5@}} \\
%     & \textbf{0.3} & \textbf{0.5} & \textbf{0.3} & \textbf{0.5} \\
%     \midrule
%     GloVe 6B  &  43.34 & 33.54 & 67.33 & 55.09\\
%     GloVe 42B  & 44.21 & 34.37  & 67.78 & 55.21 \\
%     GloVe 840B  &  44.51 & 34.87 & 65.76 & 55.24\\
%     BERT$_{Base}$  & 45.69 & 34.34  & 68.43 & 55.89 \\
%     \bottomrule
%     \end{tabular}
%     \caption{Comparison of different word embeddings on TACoS, measured by \textit{Rank 1@$\mu \in\{0.3, 0.5\}$} and \textit{Rank 5@$\mu \in\{0.3, 0.5\}$}. }
%     \label{tab:embedding}
% \end{table}

\begin{table}[t]
    \centering
    \begin{tabular}{ccccc}
    \toprule
    \multirow{2}{*}{\textbf{Methods}} & \multicolumn{4}{c}{\textbf{Rank1@}}  \\
    & \textbf{0.1} & \textbf{0.3} & \textbf{0.5} & \textbf{0.7} \\
    \midrule
    GloVe 6B  &  54.26 & 43.34 & 33.54 & 18.57\\
    GloVe 42B  & 54.74 & 44.21  & 34.37 & 20.24 \\
    GloVe 840B  &  53.11 & 44.51 & \textbf{34.87} & 19.65\\
    BERT$_{Base}$  & \textbf{57.34} & \textbf{46.26}  & 34.72 & \textbf{21.54} \\
    \bottomrule
    \end{tabular}
    \caption{Comparison of different word embeddings on TACoS, measured by \textit{Rank 1@$\mu \in\{0.1, 0.3, 0.5, 0.7\}$}. }
    \label{tab:embedding}
\end{table}

\subsubsection{Efficiency of Our RaNet}
\label{sec:params}

% \subsubsection{Efficiency of Our Relation Constructor}
% Previous methods MAN \cite{zhang2019man}, 2D-TAN \cite{zhang2020learning} also considered moment-wise temporal relations. MAN devised an iterative graph adjustment network to learn dense adjacency matrice, while our implementation is similar to 2D-TAN, which utilized a sparse sampling strategy. However, 2D-TAN simply stacked several convolution layers, which is less effective and computationally expensive, our relation constructor considers more boundary information and is implemented in a faster way. We can see in Table \ref{tab:params}, RaNet is more lightweight with only 11 M parameters compared with 92 M of 2D-TAN on ActivityNet, and the comparison of FLOPs further indicates the efficiency of our relation constructor against simple convolution layers.

% Previous methods MAN \cite{zhang2019man}, 2D-TAN \cite{zhang2020learning} also considered moment-wise temporal relations. MAN devised an iterative graph adjustment network to learn dense adjacency matrice, while our implementation is similar to 2D-TAN, which utilized a sparse sampling strategy. However, 2D-TAN simply stacked several convolution layers, which is less effective and computationally expensive, our relation constructor considers more boundary information and is implemented in a faster way. 
Both fully-connected graph neural networks and stacked convolution layers result in high computation complexity and occupy a huge number of GPU memory. With the help of a sparsely-connected graph attention module used in our Multi-choice Relation Constructor, we can capture moment-wise relations from global dependencies in a
more efficient and effective way. Table \ref{tab:params} shows the parameters and FLOPs of our model and 2D-TAN, which uses several convolution layers to capture context of adjacent moment candidates. We can see that RaNet is more lightweight with only 11 M parameters compared with 92 M of 2D-TAN on ActivityNet. Compared with RaNet, RaNet-base replaces the relation constructor with the same 2D convolutional layers as 2D-TAN. Their comparison on FLOPs further indicates the efficiency of our relation constructor against simple convolution layers.

\begin{table}[t]
    \centering
    \resizebox{\linewidth}{!}{
    \begin{tabular}{ccccc}
    \toprule
    \textbf{ } & \textbf{ } & \textbf{TACoS} & \textbf{Charades} & \textbf{ActivityNet}\\
    \midrule
    \textbf{ } & \textbf{2D-TAN} &  60.93M & 60.93M & 91.59M  \\
    \textbf{Params } & \textbf{RaNet-base} & 61.52M & 59.95M & 90.64M \\
    \textbf{ } & \textbf{RaNet} & 12.80M & 12.80M  & 10.99M \\
    \midrule
    \textbf{ } & \textbf{2D-TAN} &  2133.26G & 104.64G & 997.30G  \\
    \textbf{FLOPs } & \textbf{RaNet-base} & 2137.68G  & 104.72G & 999.54G \\
    \textbf{ } & \textbf{RaNet} & 175.36G & 4.0G  & 43.92G \\
    % \midrule
    % \textbf{ } & \textbf{2D-TAN} &  16min 50s & 6min 40s & 45min 10s  \\
    % \textbf{Traning time } & \textbf{RaNet-base} & 5min 50s$^{\ast}$ & 3min & 26min 30s \\
    % \textbf{ } & \textbf{RaNet} & 4min 50s & 1min 20s  & 4min 50s \\
    \bottomrule
    \end{tabular}
    }
    \caption{Parameters and FLOPs of our RaNet with the previous best mothod 2D-TAN, which also considers moment-level relations. M and G represent $10^{6}$ and $10^{9}$ respectively.}
    \label{tab:params}
\end{table}

\subsubsection{Qualitative Analysis}
\begin{figure}[t]
    \centering
    \includegraphics[width=1.0\linewidth]{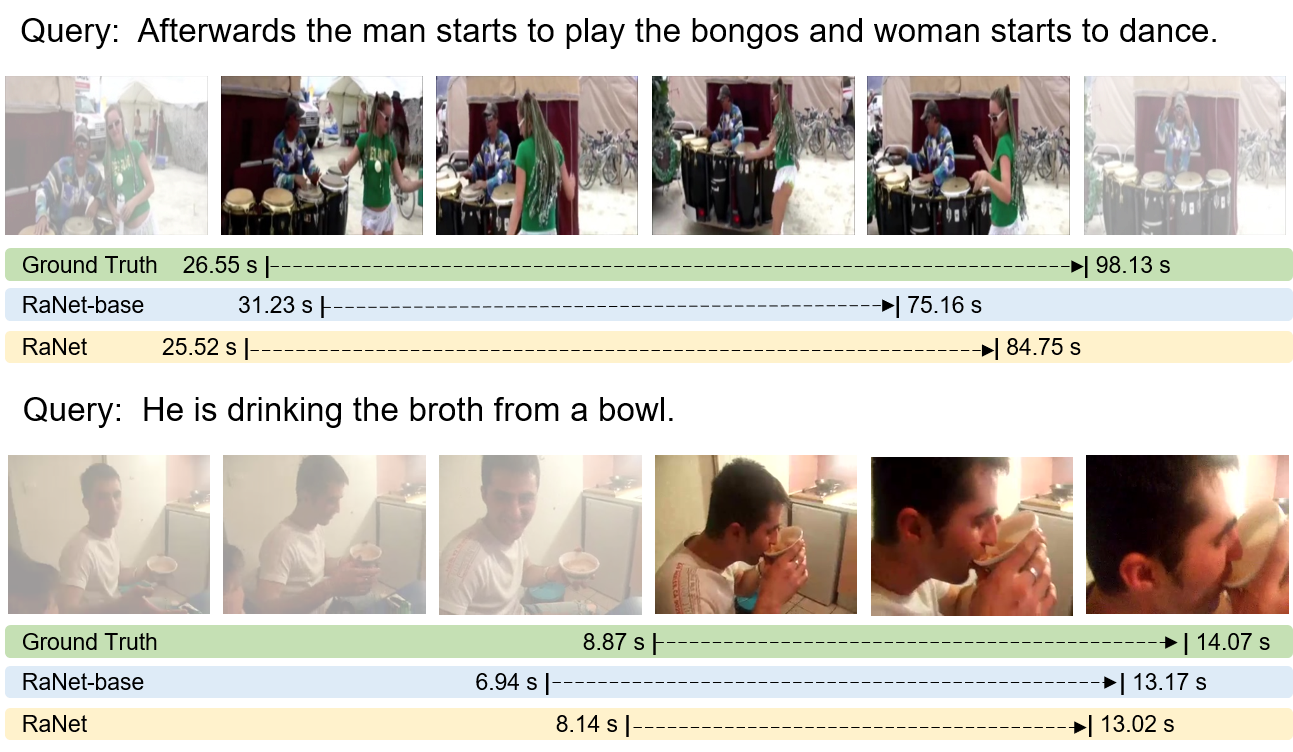}
    \caption{The qualitative results of RaNet and RaNet-base on the ActivityNet Captions dataset.}
    \label{fig:quality-analysis}
\end{figure}

We further show some examples in Figure \ref{fig:quality-analysis} from ActivityNet Captions dataset. From this comparison, we can find that predictions of our approach are closer to ground truth than our baseline model, which is the one removing $\mathbf{F}_2$ and $\mathbf{R}$ in Table \ref{tab:ablation-components}. Considering the same setting for the moment candidate, it also demonstrates the effect of our proposed modules. With the interaction and relations construction modules, our approach can select the choice of video moments matching the query sentence best. In turn, it reflects that capturing the token-aware visual representation for moment candidates and relations among candidates facilitate the net scoring candidates better.

\section{Conclusion}
In this paper, we propose a novel Relation-aware Network to address the problem of temporal language grounding in videos. We first formulate this task from the perspective of multi-choice reading comprehension. Then we propose to interact the visual and textual modalities in a coarse-and-fine fashion for token-aware and sentence-aware representation of each choice. Further, a GAT layer is introduced to mine the exhaustive relations between multi-choices for better ranking. Our model is efficient and outperforms the state-of-the-art methods on three benchmarks, i.e., ActivityNet-Captions, TACoS, and Charades-STA.
\section*{Acknowledgements}
First of all, I would like to give my heartfelt thanks to all the people who have ever helped me in this paper. The support from CloudWalk Technology Co., Ltd is gratefully acknowledged. This work was also supported by the King Abdullah University of Science and Technology (KAUST) Office of Sponsored Research through the Visual Computing Center (VCC) funding.
% National Key Research and Development Program of China (SQ2018AAA010010), Zhejiang Natural Science Foundation (LR19F020002, LZ17F020001), National Natural Science Foundation of China (61976185, 61572431), the Fundamental Research Funds for the Central Universities Chinese Knowledge Center for Engineering Sciences and Technology, and Joint Research Program of ZJU Tongdun Technology. Long Chen was supported by 2018 Zhejiang University Academic Award for Outstanding Doctoral Candidates.

% \newpage
% \bibliography{anthology,emnlp2021}
% \bibliographystyle{acl_natbib}

\bibliography{anthology,custom}
\bibliographystyle{acl_natbib}
\end{document}